\documentclass{article} 

\usepackage{nips15submit_e, times}
\usepackage{hyperref}
\usepackage{url}

\usepackage[ruled,vlined,boxed]{algorithm2e}
\usepackage{wrapfig}
\usepackage{floatflt}
\usepackage{microtype,soul}
\usepackage{graphicx}
\usepackage{multirow}
\usepackage{caption}
\usepackage{amsfonts, amssymb, amsmath, amssymb, amsthm, layout, latexsym}

\usepackage{enumitem}
\setlist{nolistsep, itemsep=0pt, parsep=0pt, leftmargin=0.7cm}
\usepackage{float}

\newtheorem{assump}{Assumption}
\newtheorem{theorem}{Theorem}
\newtheorem{lemma}{Lemma}

\title{Lower Bounds for Multi-armed Bandit with Non-equivalent Multiple Plays}

\author{
Aleksandr Vorobev\\
Yandex\\
Moscow, Russia \\
\texttt{alvor88@yandex-team.ru} \\
\And
Gleb Gusev\\
Yandex\\
Moscow, Russia \\
\texttt{gleb57@yandex-team.ru} \\
}

\nipsfinalcopy 

\begin{document}

\abovedisplayskip=.5\abovedisplayskip
\belowdisplayskip=.5\belowdisplayskip
\abovedisplayshortskip=.5\abovedisplayshortskip
\belowdisplayshortskip=.5\belowdisplayshortskip

\maketitle

\begin{abstract}
We study the stochastic multi-armed bandit problem with non-equivalent multiple plays where, at each step, an agent chooses not only a set of arms, but also their order, which influences reward distribution. In several problem formulations with different assumptions, we provide lower bounds for regret with standard asymptotics $O(\log{t})$ but novel coefficients and provide optimal algorithms, thus proving that these bounds cannot be improved.

\end{abstract}


\section{Introduction}
Multi-armed bandit (MAB) is a common model to formulate problems of finding the tradeoff between exploration
and exploitation. Its stochastic formulation with multiple plays was originally considered in~\cite{BasicPaper}. In this formulation, at each step of a game, an agent chooses $m$ arms from an arm set $A$ and observes the reward for each of them, which is a random variable whose distribution is a property of the arm. 
The agent's goal is to minimize the expected cumulative \textit{regret} over the first $T$ steps, i.e., the difference between the expected cumulative reward of the observed arms for the optimal strategy, which relies on the complete information about the reward distributions of all the arms, and the chosen strategy, which relies on the past observations only.
In the paper~\cite{BasicPaper}, theoretical analysis of the asymptotic behavior of the cumulative regret  is provided.

An important limitation of~\cite{BasicPaper} is that the rewards of the chosen arms are supposed to be independent of the order the agent put them into the set. In many applications, on the contrary, the same arm can exhibit different reward distributions at different positions.
In particular, problems of web search ranking \cite{baseline, OurPaper}, recommendations \cite{NewsRecommendation,GaussianProcesses}, and contextual advertising \cite{Advertisment, AdvertismentBandit} are often formulated as MAB problems with documents, recommended items, and ads respectively as arms.
Steps of the game correspond to the requests of users, the application (agent) chooses objects to show them in different slots (positions) of the web page, and the user's interaction with an object (which defines its reward)
clearly depends on the slot of the page the object is placed in.

Some papers studied adversarial bandit settings with non-equivalent plays~\cite{Hedge, Slate, CombinatorialBandits, Permutahedron}.
Some other studies~\cite{baseline, OurPaper, MetricSpace, GaussianProcesses} consider stochastic problem formulations and 
prove upper bounds for the regret of corresponding algorithms. 
All these algorithms follow a general scheme: they rank arms by some score which balances between exploration and exploitation, and choose the top arms for the slots in the order of the slots' importance. Thereby, these algorithms use the same
exploration rate to choose arms for different positions.
However, it follows from \cite{BasicPaper} that, in stochastic setting, even in the case of equivalent plays, an asymptotically optimal algorithm should explore only one arm at one step most part of time.


In this paper, we consider several settings of the general stochastic non-contextual MAB problem with non-equivalent multiple plays. These settings (see Section~\ref{Formalization} for description) differ by additional restrictions on the parameter space of arms and the reward distributions of their lists. These assumptions were held in many above-mentioned works and handle a variety of application tasks.
In the chosen settings, we provide lower bounds for the asymptotic behavior of the cumulative regret in Section~\ref{LowerBound} and prove their tightness under additional reliable requirements
by presenting an algorithm with the same regret asymptotic behavior. 
Importantly, the form of each lower bound gives an insight on the construction of optimal algorithms in some specific cases not covered by our algorithm.




\section{Problem Formalization}\label{Formalization}
Let us consider the following problem. 
There is a parameter space $\textbf{A}$ equipped with continuously distributed random vectors $F(\bar{a})$ with values in $\mathbb{R}^d$, densities $f(\cdot,\bar{a})$, and finite expectations $\mu(\bar{a})$ for each list $\bar{a}\in \textbf{A}^m$ of values of a fixed length $m$.
We require each component $F_i$ of $F(\bar{a})$ to be integrable: $\int_{\mathbb{R}^d}|x_i|f(x,\bar{a})\, dx<\infty$.
The case, where all distributions $f(\cdot,\bar{a})$ are discrete, can be considered as well by substituting probability functions for densities $f(\cdot,\bar{a})$ and substituting summation for integration everywhere in the paper. 

At the start, an agent is provided with the space $\textbf{A}$ and \textit{arms} $1,2,\ldots,N$, where each arm $j$ is provided with an unknown parameter $a_j\in\textbf{A}$. We denote $A:=(a_1,\ldots,a_N)$. 
At each step $t$, the agent chooses a list of different arms $\pi_t=(\pi_t(1),\ldots,\pi_t(m))$ ($\pi_t(j_1)\neq \pi_t(j_2)$ if $j_1\neq j_2$)
to fill a row of slots $S=\{1,\ldots,m\}$ with them. We denote the set of all the lists of $m$ different arms by $\Pi$.
Next, the agent observes
a realization $F_t$ of $F(\bar{a}_{\pi_t})$ ($F_t$ are independent over steps), where $\bar{a}_{\pi(t)}=(a_{\pi_t(1)},\ldots,a_{\pi_t(m)})$,
and further utilizes it for choosing lists at future steps. 
Note that $f(\cdot,\bar{a})$ can be not invariant with respect to permutations, i.e., the order of the arms in the list is important.
The agent's goal is to 
minimize the cumulative regret $\textbf{Reg}_T$ over the first $T$ steps:
$$\textbf{Reg}_T=T \max\limits_{\pi\in \Pi} \mathbb{E}R(\bar{a}_{\pi}) -\mathbb{E} \sum\nolimits_{t=1}^{T} R_t,$$
where $R(\bar{a})=U(F(\bar{a}))$, $R_t=U(F_t)$, and $U:\mathbb{R}^d \mapsto \mathbb{R}$ is the function of reward depending on the observed values.
Splitting the standard notion of an observed reward into the observed values $F$ and the reward $R$
allows to handle the case of observing only the list reward  (see \cite{CombinatorialBandits, Permutahedron}) as well as the cases when
the agent observes a contribution of each individual arm to the list reward (see, e.g., Assumption~\ref{Decomposition}) or other aspects of the interaction that can provide additional information on $a_j$, e.g., the time to the first click or the session duration in the case of web services. 
The described problem setting generalizes the one considered in~\cite{BasicPaper} to non-equivalent plays and a more general form of relation between observed values $F$ and the optimized reward $R$. 


\section{Lower Bounds for Regret}\label{LowerBound}

Before presenting each of our results, we introduce some notations and additional assumptions (on the space $(\textbf{A}, \{f(\cdot, \bar{a})\}_{\bar{a}\in \textbf{A}^m})$) this result relies on. 
The Kullback-Liebler divergence,
${I (f(\cdot),g(\cdot))=\int\nolimits_{\mathbb{R}^d} f(x) \log\frac{f(x)}{g(x)}\, dx}$,
is a widely used measure of dissimilarity between two distributions.
We denote $I (\bar{a},\bar{b})=I(f(\cdot,\bar{a}),f(\cdot,\bar{b}))$ for brevity. We assume that our space of distributions $\{f(\cdot, \bar{a})\}_{\bar{a}\in \textbf{A}^m}$ satisfies the condition $0<I(\bar{a},\bar{b})<\infty$ for any different $\bar{a},\bar{b}\in \textbf{A}^m$.
Following \cite{BasicPaper}, we consider only \textit{uniformly good strategy}, i.e., the ones with the cumulative regret of order $o(T^\alpha)$ for any $\alpha>0$ and any $A\in \textbf{A}^N$. Assume, WLOG, that each of arms $1,\ldots,m,\ldots,n$ is included in at least one optimal list (one with the highest reward expectation $\max_{\pi\in\Pi}\mathbb{E}R(\bar{a}_{\pi})$) and each of arms $n+1,\ldots,N$ is not. We call arms from these two groups \textit{relevant} and  \textit{irrelevant} respectively. 
We denote $\Pi_j:=\{\pi\in \Pi: j \in \{\pi(k)\}_{k\in S}\}$, $\bar{a}=(\bar{a}(1),\ldots,\bar{a}(m))$
and use $\bar{a}^{\{k \leftarrow a\}}$ for the list of parameter values $\bar{a}$ with $a$ substituted into the position $k$.

Our first assumption is similar to (but weaker than) the combination of Equations~2.2 and~2.4 from~\cite{BasicPaper}.
\begin{assump}\label{Condition1}
Denseness condition: 
for any list
$\bar{a}_0\in\textbf{A}^m$, slot $k$, finite set of lists $\bar{A}\subset \textbf{A}^m$, and $\rho>0$, there exists $a_0'\in \textbf{A}$ s.t.  (i)~${\mathbb{E}R(\bar{a}_0)< \mathbb{E}R(\bar{a}_0^{\{k \leftarrow a_0'\}})}$,
 (ii)~for any list $\bar{a}\in\bar{A}$ and slot $k'$ s.t. $\mathbb{E}R(\bar{a}_0^{\{k \leftarrow \bar{a}(k')\}})\neq \mathbb{E}R(\bar{a}_0)$, 
we have
${I (\bar{a},\bar{a}^{\{k' \leftarrow a_0'\}})\leq(1+\rho)I (\bar{a},\bar{a}^{\{k' \leftarrow \bar{a}_0(k)\}})}$.
\end{assump}
Assumption~\ref{Condition1} states that we can improve performance of any list by substituting such a value into an arbitrary position, which is arbitrarily ``close'' to the replaced value in terms of the reward distributions if a set of lists. This assumption holds, e.g., if ${\bf A}=\mathbb{R}$, function $I(\bar{a},\bar{b})\colon {\bf A}^{2m}\to \mathbb{R}$ is continuous and $\mathbb{E}R(\bar{a})$ is strictly monotone with respect to any $\bar{a}(k)$, $k\in S$. Denote by $N_{T}(\pi)$ the number of times list $\pi$ is used up to step $T$.  
The following lemma provides a lower bound for the regret in an implicit form and helps to obtain an explicit lower bound stated by Theorem~\ref{LowBound_Reg_Theorem} under an additional assumption. 
\begin{lemma}\label{OurTheorem}
Under Assumption~\ref{Condition1}, for any uniformly good strategy and any $A\in \textbf{A}^N$,
for any relevant arm $i\leq n$ and any irrelevant arm~$j$, the set of numbers $\{N_{T}(\pi)\}_{\pi\in \Pi_j}$ 
satisfies the following inequality:
\begin{equation}\label{OurTheorem_T}
\liminf\limits_{T\to\infty} \sum\nolimits_{\pi\in \Pi_j} \frac{\mathbb{E}N_T(\pi)}{\log{T}} I(\bar{a}_{\pi},\bar{a}_{\pi}^{\{\pi^{-1}(j) \leftarrow a_i\}}) \geq 1
\end{equation}
Consequently, there exists $x\colon \mathbb{N}\times \Pi\to \mathbb{R}_+$ such that, for all $i\leq n,j>n$, we have
$\liminf\limits_{T\to\infty} \sum\nolimits_{\pi\in \Pi_j}
x(T,\pi)
I(\bar{a}_{\pi},\bar{a}_{\pi}^{\{\pi^{-1}(j) \leftarrow a_i\}}) \geq 1$
and the cumulative regret over $T$ steps satisfies
\begin{equation}\label{Reg_LowBound_general}
\liminf\limits_{T\to\infty}\frac{\textbf{Reg}_T}{\log{T}} \geq  \liminf\limits_{T\to\infty} \sum\nolimits_{\pi\in \Pi} x(T,\pi) Reg(\pi),
\end{equation}
where  $Reg(\pi)=\max\nolimits_{\pi'\in \Pi} \mathbb{E}R(\bar{a}_{\pi'})-\mathbb{E}R(\bar{a}_{\pi})$.
\end{lemma}
This result generalizes Theorem~3.1 from \cite{BasicPaper} to two issues: (i)~a contribution of each arm to the list reward $R(\bar{a})$ may be not observed;
(ii)~the reward of the list depends on the order of the arms. 
Note that Lemma~\ref{OurTheorem} does not use any assumption on the relations between the regret distributions $f(\cdot, \bar{a})$ of different lists $\bar{a}$, e.g., overlapping in their values.
Intuitively, the less relations encoded in the space  $(\textbf{A}, \{f(\cdot, \bar{a})\}_{\bar{a}\in\textbf{A}^m})$ are, the higher
the actual regret of the optimal strategy is
(we informally call such relations {\it correlation}).
In fact, the bound from Lemma~\ref{OurTheorem_T} will be tight only in the case of ``full information'' (see, e.g., Theorem~\ref{UpperBoundMainTheorem}). One can give a formal definition for the opposite case of no information (omitted due to lack of space), when the observed rewards of one list of arms tell nothing about the reward distributions of the others.
In this case, our problem setting reduces to the standard stochastic MAB problem with single plays by considering each list as a separate arm. Within it, the tight lower bound for the regret is provided in \cite[Theorem~1]{Robbins}. Hence, we return to the setting with Assumption~\ref{Condition1}. 

An explicit bound on the regret can be found as the infimum of the right-hand side of Equation~\ref{Reg_LowBound_general} over possible functions $x(T,\pi)$. We claim, omitting a rather standard proof, that there exists $x(T,\pi)$ which provides the minimum 
and have a finite limit $y(\pi)=\lim_{T\to \infty}x(T,\pi)$ for each $\pi\in\Pi$. To find the optimal values of $y(\pi)$, we consider the Karush-Kuhn-Tucker conditions for the minimization of the right-hand side of Equation~\ref{Reg_LowBound_general} under the constraints defined by Equation~\ref{OurTheorem_T} for all $i\leq n$ and $j>n$ with $\liminf\limits_{T\to\infty}\mathbb{E}\frac{N_T(\pi)}{\log{T}}$ replaced by $y(\pi)$:
\begin{equation}\label{Karush}
\left\{
 \begin{array}{llll}
\quad \sum_{\pi\in \Pi_j} y_{\pi} I(\bar{a}_{\pi},\bar{a}_{\pi}^{\{\pi^{-1}(j) \leftarrow a_i\}})=1 & \mbox{or} & \lambda_{i,j}=0  & \mbox{for any } j>n, i\leq n\\
\quad \sum\nolimits_{k\in S} \lambda_{i,\pi(k)} I(\bar{a}_{\pi},\bar{a}_{\pi}^{\{k \leftarrow a_i\}})=Reg(\pi) & \mbox{or} & y_{\pi}=0 &  \mbox{for any } \pi\in\Pi\\ 
 \end{array}
 \right.
\end{equation}
Thus, the optimal values of $y_{\pi}$ could be found by
comparing solutions of all the linear systems over different arms $i, j$ and lists $\pi$ satisfying $\lambda_{i,j}=0$ and $y_{\pi}=0$ respectively.

However, the minimum can be found more efficiently under
the following assumption 
about decomposition of a list reward into the sum of the arms' rewards, which is almost always accepted in the literature \cite{ BasicPaper, GaussianProcesses, Slate, baseline, OurPaper}, because it is satisfied by different measures of profit for many applications.
\begin{assump}\label{Decomposition}
Decomposition condition: (i)~${R(\bar{a})=\sum_{k\in S}F(k,\bar{a}(k))}$, where $\{F(i,\bar{a}(k))\}_{k\in S}$ are independent,
(ii)~vector $F(\bar{a})$ includes $F(1,\bar{a}(1)),\ldots, F(m,\bar{a}(m))$ as its components.
\end{assump}
For example, most online measures of the web search quality cumulate some relevance gains over documents, e.g., clicks or their dwell times. 
Observability of the values
$F(1,\bar{a}(1)),\ldots,F(m,\bar{a}(m))$ (condition (ii)) is crucial for our analysis, since it allows to aggregate information about the plays of an arm in a slot regardless the arms chosen for other slots. We denote the distribution density of $F(k,a)$ by $f(\cdot,k,a)$, introduce ${I_k(a,b):=I(f(\cdot,k,a),f(\cdot,k,b))}$ and ${Reg(k,j):=\min_{\pi\in\Pi:\pi(k)=j} Reg(\pi)}$, and use $A^*_k$ for the set of arms which are placed in slot $k$ in at least one optimal list.
Assumption~\ref{Decomposition} allows us to
present the lower bound for the regret in the following simple form.
\begin{theorem}\label{LowBound_Reg_Theorem}
Under Assumptions~\ref{Condition1} and \ref{Decomposition}, for any uniformly good strategy and any $A\in \textbf{A}^N$, 
\begin{equation}\label{LowBound_Reg}
\liminf\limits_{T\to\infty}\frac{\textbf{Reg}_T}{\log{T}} \geq \sum\limits_{j>n}\max\limits_{i\leq n}\min\limits_{k\in S} \frac{Reg(k,j)}{I_k(a_j,a_i)}
\end{equation}
\end{theorem}
This result is very intuitive: the maximization means that we should distinguish an irrelevant arm $j$ from any relevant arm $i$, the minimization reflects the hope that we are able to make exploratory observations mostly in optimal slots, and the optimized component is standard.
Note that Theorem~\ref{LowBound_Reg_Theorem} improves only the representation of the lower bound given by Lemma~\ref{OurTheorem} but not the bound itself, what is impossible under Assumptions~\ref{Condition1} and \ref{Decomposition} (as we prove in Section~\ref{UpperBound}).

On the other hand, adding requirement of 
the uncorrelation between reward distributions of an arm in different slots
allows to obtain a higher lower bound in Theorem~\ref{OurTheorem_T_Cond2}. The uncorrelation combined with Assumption~\ref{Condition1} under Assumption~\ref{Decomposition} is formalized in the following assumption. 
\begin{assump}\label{No-connectedness-over-positions}
Uncorrelation-over-positions denseness condition:
for any values $a,a_0\in\textbf{A}$, slot $k$ and $\rho>0$, there exists $a_0'\in \textbf{A}$ s.t.
(i)~${\mathbb{E}F(k,a_0)<\mathbb{E}F(k,a_0')}$,
(ii)~${I_{k}(a,a_0')<(1+\rho) I_{k}(a,a_0)}$,
(iii)~for any slot $k'\neq k$, we have ${f(\cdot,k',a)=f(\cdot,k',a_0')}$.
\end{assump}
This assumption holds, e.g., if ${\bf A}=\mathbb{R}^m$, $F(k,a)=G(a^k)$ for $a=(a^1,\ldots,a^m)$, where a space of distributions $\{G(\theta)\}_{\theta\in\mathbb{R}}$ is characterized by a strictly monotone (in $\theta$) expectation function and a continuous function $I(\theta_1,\theta_2)$. 
\begin{theorem}\label{OurTheorem_T_Cond2}
Under Assumptions~\ref{Decomposition} and \ref{No-connectedness-over-positions}, for any uniformly good strategy and any $A\in \textbf{A}^N$, the number of plays $N_{T}(k,j)$ of any irrelevant arm $j>n$ in any slot $k$ during the first $T$ steps satisfies the following inequality for any arm $i\in A^*_k$:
\begin{equation}\label{PositionBound}
\liminf\limits_{T\to\infty} \frac{ \mathbb{E}N_T(k,a_j) }{\log{T}} \geq \frac{1}{I_k(a_j,a_i)},
\end{equation}
Then the cumulative regret satisfies the following lower bound:
\begin{equation}\label{LowerBound_NoConnectedness}
\liminf\limits_{T\to\infty}\frac{\textbf{Reg}_T}{\log{T}} \geq \sum\limits_{j>n,k\in S}\max_{i\in A^*_k} \frac{Reg(k,a_j)}{I_k(a_j,a_i)}
\end{equation}
\end{theorem}
Naturally, in order to distinguish the arm $j$ from the arm $i$ in the slot $k$, we need to play it in this slot the same number of times as in the standard SMAB problem with one play and the optimal arm $i$. Though reward distributions of an object in different slots seem to be dependent in practice, it may be of use for constructing a strategy to treat them as independent if the dependence is difficult to be inferred. As an example, one can consider a project with various tasks requiring different competencies and to be assigned to different workers from a big set of candidates, e.g., a football match, where a manager chooses players for different positions.
We also note that $\max$ in Equation~\ref{LowerBound_NoConnectedness} disappears if there is the only optimal list.
Now we prove our claims.

\textbf{Proof of Lemma~\ref{OurTheorem}.}
At the first step of our proof, we use the change of measure technique, like \cite{BasicPaper} does, and prove that, for any irrelevant arm $j>n$,
the vector of numbers $\{N_T(\pi)/\log{T}\}_{\pi\in\Pi_j}$, with high probability, lays outside of some $|\Pi_j|$-dimensional cuboids of the form ${\{\{x(\pi)\}_{\pi\in\Pi_j}: 0\leq x_{\pi}< c(\pi)\}}$. At the second step, which is completely novel and crucial for the new issues,  we aggregate these estimates to show that this vector is outside of a sequence of simplexes what in the limit provides Equation~\ref{OurTheorem_T}.

\textbf{Step 1}. 
Consider any optimal arm list $\pi_0$, any arm $i\in \pi_0(S)$, and any irrelevant arm $j>n$. According to Assumption~\ref{Condition1} applied to the list $\pi_0$, the slot $\pi_0^{-1}(i)$ and the set of lists $\Pi_j$, for a fixed $\rho>0$, we can choose a value $a^* \in \textbf{A}$ such that
\begin{equation}\label{ArmChange}
(i)~\mathbb{E}R(\bar{a}_{\pi_0}^{\{\pi_0^{-1}(i)\gets a^*\}}> \mathbb{E}R(\bar{a}_{\pi_0}),
(ii)~(1+\rho)I(\bar{a}_{\pi},\bar{a}_{\pi}^{\{\pi^{-1}(j) \gets a_i\}})>I(\bar{a}_{\pi},\bar{a}_{\pi}^*) \ \ \forall \pi\in \Pi_j,
\end{equation}
where we denote $\bar{a}_{\pi}^*:=\bar{a}_{\pi}^{\{\pi^{-1}(j) \gets a^*\}}$ for $\pi\in \Pi_j$. We use the ``alternative'' parameter values ${A^*=\{a_1,\ldots,a_{j-1}, a^*,a_{j+1},\ldots,a_N\}}$ to prove the following statement.
\begin{lemma}\label{LemmaCuboid} Consider any $\overline{c}=\{c({\pi})\}_{\pi\in\Pi_j}$ satisfying $\sum\limits_{\pi\in\Pi_j} c(\pi)I(\bar{a}_{\pi},\bar{a}^*_{\pi})=\delta < \frac{1}{1+\rho}$. We have
\begin{equation}\label{lim=zero}
\lim\limits_{T\to \infty} P_A\left(\prod\nolimits_{\pi\in\Pi_j} \{ N_T(\pi)/\log{T}<c(\pi) \}\right) =0.
\end{equation}
\end{lemma}
The proof is based on the log odds ratio of the likelihood of the rewards $R_{1}(\bar{a}_{\pi}),\ldots,R_{t}(\bar{a}_{\pi})$ (observed at $t$ plays of a list $\pi$) under the parameter values $\bar{a}_1$ and $\bar{a}_2$: ${L_{t,\pi}(\bar{a}_1,\bar{a}_2)=\sum\nolimits_{\tau=1}^t \log \frac{f(R_{\tau}(\bar{a}_{\pi}),\bar{a}_1)}{f(R_{\tau}(\bar{a}_{\pi}),\bar{a}_2)}}$.
By the strong law of large numbers, we have
$$
{I(\bar{a}_{\pi},\bar{a}_{\pi}^*)=\mathbb{E}_A\left (\log
\frac{f(R(\bar{a}_{\pi}),\bar{a}_{\pi})}{f(R(\bar{a}_{\pi}),\bar{a}_{\pi}^*)}\right ) = \lim\limits_{t\to \infty} \frac{L_{t,\pi}(\bar{a}_{\pi},\bar{a}_{\pi}^*)}{t}=\lim\limits_{t\to \infty} \frac{\max\limits_{\tau\leq t} L_{\tau,\pi}(\bar{a}_{\pi},\bar{a}_{\pi}^*)}{t} \mbox{ \,\, $P_A$-a.s.}}
$$
Consequently, for $\pi\in \Pi_j$ and events
${B_{\tau,T}^{\pi,c}:=\{L_{\tau,\pi}(\bar{a}_{\pi},\bar{a}^*_{\pi})\leq (1+\rho)I (\bar{a}_{\pi},\bar{a}^*_{\pi})c\log{T} \}}$, we have
\begin{equation}\label{asympt_2}
\lim\limits_{T \to\infty} P_A(\prod\nolimits_{\tau<c\log{T}}B_{\tau,T}^{\pi,c})=1.
\end{equation}
Using Equation~\ref{asympt_2}, we obtain Lemma~\ref{LemmaCuboid} in the following way:
\begin{multline}
\lim\limits_{T\to \infty} P_A\left(\prod\limits_{\pi\in\Pi_j} \{ \frac{N_T(\pi)}{\log{T}}<c(\pi) \}\right)\leq \lim\limits_{T\to \infty} \bigg[P_A\left( \prod\limits_{\pi\in\Pi_j}\left(\{ \frac{N_T(\pi)}{\log{T}}<c(\pi)\} \prod_{\tau<c(\pi)\log{T}} B_{\tau,T}^{\pi,c(\pi)}\right)\right)+\\
+1-P(\prod\limits_{\pi\in\Pi_j}\prod_{\tau<c(\pi)\log{T}} B_{\tau,T}^{\pi,c(\pi)})\bigg]\leq
 \lim\limits_{T\to \infty} P_A\left(\prod\limits_{\pi\in\Pi_j}\{ \frac{N_T(\pi)}{\log{T}}<c(\pi)\} B_{N_T(\pi),T}^{\pi,c(\pi)}\right) =0.
\end{multline}
To prove the last equality, we introduce, for any $\overline{\tau}=\{\tau(\pi)\}_{\pi\in \Pi_j}$ s.t. $\tau(\pi)<c(\pi)$, event $S_j(T, \overline{c}, \overline{\tau}):=\prod\nolimits_{\pi\in\Pi_j}(\{ N_T(\pi)=\tau(\pi)\} B_{N_T(\pi),T}^{\pi,c(\pi)})$ and find: \begin{multline}\label{KeyEquation}
P_{A^*} (S_j(T, \overline{c}, \overline{\tau}))=
\int\textbf{1}\{S_j(T, \overline{c}, \overline{\tau})\}\, dP_{A^*} =\int\textbf{1}\{S_j(T, \overline{c}, \overline{\tau})\} e^{-\sum\nolimits_{\pi\in\Pi_j} L_{\tau(\pi),\pi}(\bar{a}_{\pi},\bar{a}^*_{\pi})}\, dP_{A} \geq\\
\geq T^{-(1+\rho)\sum\nolimits_{\pi\in\Pi_j} c(\pi)I(\bar{a}_{\pi}, \bar{a}^*_{\pi})} P_{A}\left(S_j(T, \overline{c}, \overline{\tau})\right)\geq
T^{-\delta (1+\rho)} P_{A}\left(S_j(T, \overline{c}, \overline{\tau})\right),
\end{multline}
where the second equality uses the change of measure, which concerns only arm $j$, and the thirst inequality is based on the definition of $B_{N_T(\pi),T}^{\pi,c(\pi)}$.
Since
$\prod\nolimits_{\pi\in\Pi_j}\{ N_T(\pi)<c(\pi)\log{T}\} B_{N_T(\pi),T}^{\pi,c(\pi)}=
\bigcup\limits_{\overline{\pi}:\tau(\pi)<c(\pi)\log{T}}\prod\limits_{\pi\in\Pi_j} S_j(T, \overline{c}, \overline{\tau})$,
where united sets are disjoint, we obtain from Equation~\ref{KeyEquation}
\begin{multline}\label{asympt_3}
P_A\left(\prod\limits_{\pi\in\Pi_j} \{ \frac{N_T(\pi)}{\log{T}}<c(\pi)\} B_{N_T(\pi),T}^{\pi,c(\pi)} \right) \leq
T^{\delta (1+\rho)} P_{A^*}\left(\prod\limits_{\pi\in\Pi_j} \{ \frac{N_T(\pi)}{\log{T}}<c(\pi)\} B_{N_T(\pi),T}^{\pi,c(\pi)}\right)
\end{multline}
Equation~\ref{ArmChange} (i) implies that, under $A^*$, any optimal list $\pi_{opt}$ belongs to $\Pi_j$. Since the strategy is uniformly good, we also have
$P_{A^*} \{\frac{N_T(\pi_{opt})}{\log{T}}<c\}\leq   \frac{ \mathbb{E}_{A^*}(T-N_T(\pi_{opt}))}{T-c \log{T}}=\frac{o(T^{\alpha})}{T-c \log{T}}=o(T^{\alpha-1})$ for any $c>0$. Therefore, Equation~\ref{asympt_3} implies that its left-hand side is $o(T^{\alpha-1+\delta (1+\rho)})=o(1)$, if we choose $\alpha\in (0,1-\delta (1+\rho))$.

\textbf{Step 2}. 
Fix any $\epsilon>0$ and choose $\rho>0$ such that $\frac{1}{(1+2\rho)(1+\rho)}>1-\epsilon$. We obtain Equation~\ref{OurTheorem_T} from
$$\mathbb{E}\frac{\sum\limits_{\pi\in \Pi_j} N_T(\pi) I(\bar{a}_{\pi},\bar{a}_{\pi}^{\{\pi^{-1}(j) \gets a_i\}})}{\log{T}} \geq
P_A\left( \frac{\sum\limits_{\pi\in \Pi_j} N_T(\pi) I(\bar{a}_{\pi},\bar{a}_{\pi}^{\{\pi^{-1}(j) \gets a_i\}})}{\log{T}} \geq 1-\epsilon\right)(1-\epsilon)\xrightarrow[T\to \infty]{} 1-\epsilon.$$
This convergence is equivalent to
\begin{equation}\label{SimplexConvergence}
\lim\nolimits_{T\to \infty} P_A\left(\left\{N_T(\pi)/\log{T}\right\}_{\pi\in\Pi_j} \in S_{1-\epsilon}\right) =0
\end{equation}
for the simplex
$S_{1-\epsilon}=\left\{\{x(\pi)\}_{\pi\in \Pi_j}: \sum\nolimits_{\pi\in \Pi_j} x(\pi) I(\bar{a}_{\pi}, \bar{a}_{\pi}^{\{\pi^{-1}(j) \gets a_i\}})  <1-\epsilon, x(\pi)\geq 0 \right\}$. Note that it
can be covered by a finite union of cuboids
$C_{\{c(\pi)\}_{\pi\in \Pi_j}}=\left\{\{x(\pi)\}_{\pi\in \Pi_j}: 0\leq x(\pi)<c(\pi) \right\}$
 contained in $S_{\frac{1}{(1+2\rho)(1+\rho)}}$, i.e., satisfying
$\sum\nolimits_{\pi\in \Pi_j} c(\pi) I(\bar{a}_{\pi}, \bar{a}_{\pi}^{\{\pi^{-1}(j) \gets a_i\}})  <\frac{1}{(1+2\rho)(1+\rho)}$.
Due to Equation~\ref{ArmChange}, the latter condition implies
$\sum\nolimits_{\pi\in \Pi_j} c(\pi) I(\bar{a}_{\pi}, \bar{a}^*_{\pi})  <\frac{1}{1+2\rho}$.
Then, from Equation~\ref{lim=zero}, for each of these cuboids,
$\lim\nolimits_{T\to \infty} P_A\left(\left\{N_T(\pi)/\log{T}\right\}_{\pi\in\Pi_j} \in C_{\{c(\pi)\}_{\pi\in \Pi_j}}\right) =0$ what implies Equation~\ref{SimplexConvergence}.

Finally, 
Equation 1.1 from~\cite{Robbins} yields 
${\liminf\limits_{T\to\infty}\frac{\textbf{Reg}_T}{\log{T}} = \liminf\limits_{T\to\infty}\sum\limits_{\pi\in \Pi} \mathbb{E}N_T(\pi) Reg(\pi)/\log{T}}$,
and we obtain Equation~\ref{Reg_LowBound_general} by minimizing this expression over $\{\mathbb{E}N_T(\pi)\}_{T\in\mathbb{N},\pi\in \Pi}$ satisfying Equation~\ref{OurTheorem_T} for all $i\leq n,j>n$. Existence of an optimal function $x(T,\pi)$ is discussed above, before Equation~\ref{Karush}.
\qed

\textbf{Proof of Theorem~\ref{LowBound_Reg_Theorem}}.
Under Assumption~\ref{Decomposition}, 
we have
$f((x^1,\ldots,x^m),\bar{a})=f(x^1,\bar{a}(1))\ldots f(x^m,\bar{a}(m))$, and, thus, for any list $\bar{a}\in \textbf{A}^m$, slot $k$ and value $a^*\in \textbf{A}$,
\begin{multline}
I(\bar{a},\bar{a}^{\{k \gets a^*\}})=\int\limits_{\mathbb{R}^m} f(x^1,\bar{a}(1))\ldots f(x^m,\bar{a}(m))\log{\frac{f(x^k,\bar{a}(k))}{f(x^k,a^*)}}dx^1...dx^m=I_k(\bar{a}(k),a^*)
\end{multline}
Then, we can rewrite Equation~\ref{OurTheorem_T} as follows:
\begin{equation}\label{LowBound_T_m}
\liminf\nolimits_{T\to\infty} \mathbb{E}\sum\nolimits_{k\in S} N_T(k,j) I_k(a_j,a_i)/\log{T} \geq 1
\end{equation}
For further simplification of these restrictions, we utilize the following combinatorial lemma which claims that, in order to minimize regret under the fixed values of $\{N_T(k,j)\}_{j>n,k\in S}$, a strategy should not observe several arms $j>n$ at one step.
\begin{lemma}\label{CombinatorialLemma}
There are $m$ slots and $m$ objects with some reward $r(k,j)$ corresponding to an object $j$ put in a slot $k$. Let consider $t\leq m$ steps and a subset of different slots $\{k_1,\ldots,k_t\}$. Assume we should close each of these slots at exactly one step. After it, at each step, we choose such a combination of different objects to put them in open slots (only one object in one slot) that maximizes the cumulative reward on this step. Then, one of the ways to reach the maximum cumulative reward over all the steps is to close one slot per step. 
\end{lemma}
\textbf{Proof sketch.}
The idea of the proof is that, when closing just one slot at each step, we can repeat any combination $\{N_T(k,j)\}_{k=1,\ldots,m, j=1,\ldots,m}$ which can be reached by any other strategy. We drop the accurate proof due to its technical nature.
\qed

Then, while considering only rational strategies from Lemma \ref{CombinatorialLemma}, each play of the arm $j$ in the slot $k$ corresponds to a step with regret not less than $Reg(k,j)$, what leads to the following estimate: 
\begin{multline*}\label{Reg_LowBound_narrow}
\liminf\limits_{T\to\infty}\textbf{Reg}_T/\log{T}\geq \liminf\nolimits_{T\to\infty}\sum\nolimits_{j>n} \sum\nolimits_{k\in S} N_T(k,j) Reg(k,j)/\log{T}=\\
(\forall \{i_j\}_{j>n}, i_j\leq n)\ \  =\liminf\limits_{T\to\infty}\sum\nolimits_{j>n} \sum\nolimits_{k\in S} \frac{N_T(k,j)I_k(a_j,a_{i_j})}{\log{T}} \frac{Reg(k,j)}{I_k(a_j,a_{i_j})}\geq\\
\geq\liminf\limits_{T\to\infty}\sum\limits_{j>n}\left[\min\limits_{k\in S} \frac{Reg(k,j)}{I_k(a_j,a_{i_j})} \cdot  \sum\limits_{k\in S} \frac{N_T(k,j)I_k(a_j,a_{i_j})}{\log{T}} \right]\geq
 \sum\limits_{j>n}\min\limits_{k\in S} \frac{Reg(k,j)}{I_k(a_j,a_{i_j})}, 
\end{multline*}
where the last inequality follows from Equation~\ref{LowBound_T_m}. Taking maximum over all possible sets $\{i_j\}_{j>n}$ yields Equation~\ref{LowBound_Reg}.
\qed

\textbf{Proof of Theorem~\ref{OurTheorem_T_Cond2}.}
We describe a modification of Step 1 of the proof of Lemma~\ref{OurTheorem} which proves the current theorem. Given an irrelevant arm $j>n$ and an optimal list $\pi_0$ with an arm $i$ in a slot $k$, according to Assumption~\ref{No-connectedness-over-positions}, we can choose such a value $a^*\in\textbf{A}$ that
\begin{equation}\label{ArmChange_2}
\mathbb{E}F(k,a_i)<\mathbb{E}F(k,a^*), I_{k}(a_j,a^*)<(1+\rho) I_{k}(a_j,a_i), 
\forall k'\neq k \ f(\cdot,k',a_j)=f(\cdot,k',a^*)
\end{equation}
Then, in the case of the arm parameters ${A^*=\{a_1,\ldots,a_{j-1}, a^*,a_{j+1},\ldots,a_N\}}$, the list $\pi_0$ is optimal and, since the probability to observe a fixed reward at some step in some slot differs under measures $P_A$ and $P_{A^*}$ only if the slot is $k$ with arm $j$ in it at this step, 
we can estimate each value $N_T(k,j),k\in S,$ separately. Indeed, 
by choosing ${c(\pi)\leq\frac{1}{(1+2\rho)I_k(a_j,a^*)}}$ for each list $\pi$ with $\pi(k)=j$ and putting $c(\pi)=+\infty$ for other lists, we obtain estimates analogous to Equations~\ref{KeyEquation}--\ref{lim=zero} resulting in 
$\lim\nolimits_{T\to \infty} P_A\left(  N_T(k,j)<\frac{\log{T}}{(1+2\rho) I_k(a_j,a^*)} \right)=0$ at the end of Step 1 of the proof of Lemma~\ref{OurTheorem}. Applying the first inequality from Equation~\ref{ArmChange_2} and letting $\rho\to 0$ yields Equation~\ref{PositionBound}. Maximizing its right-hand side over $i\in A^*_k$ leads to Equation~\ref{LowerBound_NoConnectedness}.
\qed


\section{Asymptotically Optimal Algorithm}\label{UpperBound}
In this section, we construct algorithms with asymptotically optimal regret reaching lower bounds from Lemma~\ref{OurTheorem} and Theorems~\ref{LowBound_Reg_Theorem} and \ref{OurTheorem_T_Cond2}. 
In our construction, we rely on the algorithm proposed in \cite{BasicPaper} under Assumptions~\ref{Condition1} and \ref{Decomposition} with additional constraint ${f(\cdot,i,a)=f(\cdot,a)}$ and modify it to handle the case of non-equivalent plays. 
First, we supplement the setting of Theorem~\ref{LowBound_Reg_Theorem} with the following assumption under which we are able to present an optimal algorithm.
\begin{assump}\label{FactorizationAssumption} Factorization condition: 
the arm reward has a form $F(k,a)=p(k)r(a)$, 
where $p(k)$ is a Bernoulli random variable with a parameter dependent on $k$ only, $r(a)$ is a random variable with distribution dependent on $a$ only. Besides, values of $p(k), k\in S,$ are components of $F(\bar{a})$. 
\end{assump}
This factorization model is often used in different applied problems, e.g., it corresponds to the examination hypothesis~\cite{ExaminationHypothesis} underlying different models of user behavior on the web search result page. Under this hypothesis, the variable $p(k)$ indicates whether the user examined the document, and $r(a)$ measures the user satisfaction with it. One of possible ways to observe these values is to consider a click or a click on a lower position as a fact of examination and a satisfied click (e.g., one with a long enough dwell time or the last click in the session \cite{30sec_1,30sec_2}) as a fact of user satisfaction.
More general but similar factorization assumption was also considered in \cite{GaussianProcesses}.

 We denote $\mathbb{E}p(k)=p_k, \mathbb{E}r(a_j)=\mu_j$ and assume, WLOG, that
$\mu_{a_1}\geq\ldots\geq\mu_{a_{l}}>\mu_{a_{l+1}}=\ldots=\mu_{a_m}=\ldots=\mu_{a_{n}}>\mu_{a_{n+1}}\geq\ldots\geq\mu_{a_N}$
 and the slots $1,\ldots,m$ are ordered by decreasing $p_k$. Below we assume that the agent knows this order. Otherwise, it could sort the slots by current empirical estimates of $p_k$. Errors of these estimates will not 
influence on the asymptotic behavior of the regret, due to the exponential convergence rate of the mean estimate provided by the Chernoff-Hoeffding bound: for iid random variables $x_1,\ldots,x_n$ with values in $[0,1]$ and for any $\epsilon>0$, $P((x_1+\ldots+x_K)/K-\mathbb{E}x_i<-\epsilon)\leq e^{-2K\epsilon^2}$. 

Due to Assumption~\ref{FactorizationAssumption}, the value of $r(a)$ is observed only if $p(k)=1$ for the corresponding slot. Further, when it is observed, its distribution does not depend on the slot. Then, we define arm-dedicated statistics $\mu_{j,t}$ and $U_{j,t}$ from \cite{BasicPaper} which, in our case, are based not on all the plays of the arm $j$ but only on all the observations of $r(a_j)$.
First one $\mu_{j,t}$ estimates expectation $\mu_j$ of $r(a_j)$: $\mu_{j,t}= \sum\nolimits_{i=1}^{N^*_t(j)} r_i(a_j)/N^*_t(j)$, where $N^*_t(j)$ is the number of observations of $r(a_j)$ during the first $t$ steps and $r_i(a_j)$ is the $i$-th observed value.
The second statistics $U_{j,t}=g_{t,N^*_t(j)}(r_1(a_j),\ldots,r_{N^*_t(j)})$ (see definition of $g_{t,s}(Y_1,\ldots,Y_s)$ in Section IV of \cite{BasicPaper}; we define $Y_i$ as an observation of $r(a)$ for some $a$) is a kind of an upper confidence bound used in different MAB algorithms, e.g., UCB-1 \cite{UCB} Bayesian-UCB \cite{Bayesian-UCB} 
and is constructed to satisfy the asymptotic properties proved in Theorem 4.2 in \cite{BasicPaper} 
under the following assumption. 
\begin{assump}\label{ThetaAssumption}
(i) The space of reward distributions can be parametrized by 
$\theta\in\mathbb{R}$, i.e., ${f(\cdot,k,a)=f(\cdot,\theta)}$, in such a way that $\log{f(x,\theta)}$ is concave in $\theta$ for each $x$. (ii)$\int x^2 f(x,\theta) \, dx<\infty$.
\end{assump}
Based on the statistics $\mu_{j,t}$ and $U_{j,t}$, we describe an asymptotically optimal algorithm under Assumptions~\ref{Condition1}, \ref{Decomposition}, \ref{FactorizationAssumption} and \ref{ThetaAssumption} in Algorithm~\ref{Algorithm}. Given values of the statistics, it chooses $m$ arms to be observed as the algorithm from \cite{BasicPaper} does it and ranks them by decreasing $\mu_{j,t}$. Theorem~\ref{UpperBoundMainTheorem} states its optimality.
\begin{algorithm}[h]
\SetAlgoNoEnd
\LinesNumbered
\SetAlgoLined
\KwData{$m$, space $(\textbf{A},\{f(\cdot,\bar{a})\}_{\bar{a}\in \textbf{A}^m})$, arm parameters $A$, slots $S$;} 
Make $m$ observations of $r_j$ (with $p(k)=1$) for each arm $j$ (in any slots); $t_0\gets \#$ of steps for it;\\
Choose $\delta\in(0,p_m/(2N^2))$;\\
\For{$t=t_0$ \KwTo $T$}{$j^*\leftarrow t\%N$ \tcp*[r]{\tiny{choose an arm uniformly over steps; $x\%y$ is a remainder of division of $x$ by $y$;}}
$G \leftarrow \varnothing$; 
\For{$j=1$ \KwTo $N$}{\lIf{$N^*_t(j)>\delta t$}{$G \leftarrow G\cup\{j\}$}}
\lIf{$|G|<m$}{Add different $(m-|G|)$ arms to $G$ randomly} \mbox{}\\
$\pi'_t\gets$ list of top-$m$ arms from $G$ by decreasing $\mu_{j,t}$; \mbox{}\\ 
\eIf{$j^*\in\pi'_t(S)$}{Show $\pi_t:=\pi'_t$;}{\leIf{ $U_{j^*,t}<\mu_{\pi'_t(m),t}$}{Show $\pi_t:=\pi'_t$;}{Show $\pi_t:=(\pi'_t(1),\ldots,\pi'_t(m-1),j^*)$}Observe user feedback $F(\bar{a}_{\pi_t})$;}}
\KwResult{ Arm list at each step: $\{\pi_t\}_{t=1,\ldots,T}$} 
\caption{Asymptotically optimal bandit algorithm under Assumptions~\ref{Condition1}, \ref{Decomposition}, \ref{FactorizationAssumption} and \ref{ThetaAssumption}}\label{Algorithm}
\end{algorithm}
\begin{theorem}\label{UpperBoundMainTheorem}
Algorithm \ref{Algorithm} is asymptotically optimal under Assumptions~\ref{Condition1}, \ref{Decomposition}, \ref{FactorizationAssumption} and \ref{ThetaAssumption}, i.e., the asymptotics of its regret coincides with the lower bound from Theorem~\ref{LowBound_Reg_Theorem}:
$\liminf\limits_{t\to\infty}\frac{\textbf{Reg}_t}{\log{t}} = \sum\nolimits_{j>n} \frac{\mu_{a_m}-\mu_{a_j}}{p_m I(a_j,a_m)}$.
\end{theorem}

\textbf{Proof.}
The following estimates show that, under Assumption~\ref{FactorizationAssumption}, the latter asymptotics corresponds to the lower bound: 
$I_k(a_j,a_{i})=p_k I(a_j,a_{i})\geq p_k I(a_j,a_{m})$, $\frac{Reg(k,a_j)}{p_k}= \frac{(\mu_{a_k}-\mu_{a_j})(p_k-p_{k+1})+\ldots+(\mu_{a_{m-1}}-\mu_{a_j})(p_{m-1}-p_{m})+(\mu_{a_m}-\mu_{a_j})p_m}{p_k}\geq
\frac{(\mu_{a_m}-\mu_{a_j})((p_k-p_{k+1})+\ldots+(p_{m-1}-p_{m})+p_m)}{p_k}= \mu_{a_m}-\mu_{a_j}$.
Further proof 
differs from that of Theorem 5.1 from~\cite{BasicPaper} by the two issues: (i) an optimal arm list with a suboptimal order of arms provides the zero regret in the original case and a non-zero one in our case; (ii) in our case, a play of an arm $j$ does not necessarily provide an observation of $r(a_j)$.
Our proof consists of the following steps corresponding to steps from~\cite{BasicPaper}.
\begin{itemize} 
\item \textit{Step A.} For each relevant arm $i\leq l$, we have $\mathbb{E}N_T(i,i)=T-o(\log{T})$.
\item \textit{Step B.} $\mathbb{E}B_T=T-o(\log{T})$, where $B_T=\#\{t\leq T\ |\ \pi'_{t} \mbox{ consists only of arms } j\leq n \}$, where $\pi'_t$ is defined in line~8 of Algorithm~\ref{Algorithm}.
\item \textit{Step C.} 
For any $j>n$ and $\rho>0$ there exists $\epsilon>0$ such that $\mathbb{E}S_T(j)\leq \frac{1+\rho+o(1)}{I_m(a_j,a_m)}\log{T}$, where\\
$S_T(j)=\#\{t\leq T\ |\  \pi'_t(i)=i \ \forall i\leq l,\ |\mu_{i,t}-\mu_{i}|<\epsilon\ \forall i\leq n, \pi'_t \mbox{ consists only of arms } {i\leq n} \mbox{ and } r(a_j) \mbox{ is observed at step } t\}$.
\end{itemize}
Now we explain how Steps A, B and C are combined to yield Theorem~\ref{UpperBoundMainTheorem}.  Let consider the following particular case of the Chernoff-Hoeffding bound for independent observations of the indicator of an $r(a_j)$ observation given an arm $j$ is played at a particular step:
\begin{equation}\label{Chernoff}
P(N_t^*(j)<N_t(j) p_m/2)\leq e^{-N_t(j) p_m^2/2}, 
\end{equation}
where $N_t(j)$ is the number of plays of the arm $j$ during the first $t$ steps. 
Along with the condition $\delta<p_m/(2N^2)$, this estimate implies that both 
the expected number of steps of Algorithm~\ref{Algorithm} with active line~7 and the expected number of steps in line~1 are finite. 
Combined with Steps A and B, this provides that the cumulative regret of Algorithm~\ref{Algorithm} is of order $o(\log{T})$, except for steps counted by $S_T(j), j>n$ at Step C. The regret at these steps is at most 
$$\sum\nolimits_{j>n} \frac{(\mu_{a_m}-\mu_{a_j})(1+\rho+o(1))}{I_m(a_j,a_m)}\log{T}=\sum\nolimits_{j>n} \frac{(\mu_{a_m}-\mu_{a_j})(1+\rho)}{p_m I(a_j,a_m)}\log{T}+o(\log{T})$$
Letting $\rho\to 0$ concludes the proof.
\qed

\textbf{Proof of Step A.}
Choose $c>(N+1)(1-2N^2\delta/p_m)^{-1}$ to provide $[(c^r-c^{r-1})/N]>2N\delta c^r/p_m$ for $r\in \mathbb{N}$ and choose $\epsilon<\min{\{(\mu_{a_i}-\mu_{a_j})/2, i<j\leq l; (\mu_{a_{l}}-\mu_{a_m})/2;(\mu_{a_{m}}-\mu_{a_{n+1}})/2\}}$. Lemmas~5.1 and 5.2 and their proofs could be transfered from \cite{BasicPaper} to our case without any changes. Now we change Lemma~5.3 from \cite{BasicPaper} by the following extended analysis. By Lemma~5.2, on $A_r B_r$, any arm $i\leq l$ satisfies $N_t(i)\geq [(c^r-c^{r-1})/N]>2\delta t/p_m$ for any step $t\in [c^r,c^{r+1}]$ and $r\geq r^*$ for some $r^*$. Then, we estimate $N_t^*(i)$ 
by Equation~\ref{Chernoff}: 
${P(N_t^*(i)<\delta t \mid A_r B_r)\leq e^{-\delta t p_m}}$. In combination with Lemma~5.1, it implies $P(C_r)=1-o(c^{-r})$ for 
$C_r=\prod_{i\leq l}\{T_t^*(i)\geq\delta t\} A_r B_r$.
Further, at step $t$, on $C_r$,  each arm $i\leq l$ is included in $\pi'_t$ in line~6 and, moreover, $\pi'_t(i)=i$, i.e., it is played at its optimal position, since on $A_r$ Algorithm~\ref{Algorithm} sorts $\pi'_t$ perfectly in line 8. Finally, we can estimate \\
$\mathbb{E}N_t(i,i)\geq\sum\nolimits_{r=r_0}^{[\log{t}]}\sum\limits_{c^r\leq t<\min\{c^{r+1},t\}} P(C_r)=\sum\nolimits_{r=r_0}^{[\log{t}]} (\min\{c^{r+1},t\}-c^r-o(1))=t-o(\ln{t})$.
\qed

\textbf{Proof of Step B.}
Again, we transfer from \cite{BasicPaper} without changes Lemmas~5.1.B and 5.2.B and the claim proved just after the proof of Lemma~5.2.B. Then, we apply the same trick as at Step A.
\qed

\textbf{Proof of Step C.}
Note that each observation of $r(a_j)$ counted by $S_t(j)$ occurs in the slot $m$. Then, we transfer the proof of Step C from \cite{BasicPaper} to our case by changing the notion of a play of the arm $j$ by the notion of an observation of $r(a_j)$.
\qed

Thus, we proved the tightness of the lower bound for the asymptotic behavior of regret provided by Lemma~\ref{OurTheorem} and Theorem~\ref{LowBound_Reg_Theorem}. 
Under Assumptions~\ref{Decomposition}, \ref{No-connectedness-over-positions} and \ref{ThetaAssumption}, a construction of an optimal algorithm reaching the lower bound from Theorem~\ref{OurTheorem_T_Cond2} could be similar. Specificity is that 
(i) the agent should maintain statistics $\mu_{k,j,t}$ and $U_{k,j,t}$ for each pair of a slot $k$ and an arm $j$ and (ii) if, at some step, the agent decides to substitute the arm from the special arm-slot pair for the arm $j$ greedily chosen for this slot, it should find a greedy-optimal combination of arms for other slots again since it may now include $j$.



\section{Conclusion}\label{Conclusion}
In this paper, we systematically studied the stochastic non-contextual multi-armed bandit problem with non-equivalent multiple plays. We considered some of the most interesting and, at the same time, quite general problem settings which are covered by our formulation and handle many applied problems. For them, we provided lower bounds for asymptotic behavior of the regret and proved tightness of these bounds. We believe that this work could be a basis both for finding theoretically optimal algorithms 
in more specific cases of our problem settings and for future development of applied algorithms.




\newpage
\bibliographystyle{abbrv}
\bibliography{Literature}

\appendix




\end{document}